\documentclass[preprint,review,number,sort&compress]{elsarticle}

\usepackage{algorithmic}
\usepackage{multirow}
\usepackage{booktabs}
\usepackage{caption}
\usepackage{graphicx}
\usepackage{subfigure}
\usepackage{ulem}
\usepackage{bm}
\usepackage{enumerate}
\usepackage{amsfonts,amssymb}
\usepackage[title]{appendix}
\usepackage{hyperref}
\usepackage{threeparttable}
\usepackage{amsmath}
\usepackage{color}

\begin{document}
	
	\begin{frontmatter}

		\title{A Deep Spatio-Temporal Architecture for Dynamic Effective Connectivity Network Analysis Based on Dynamic Causal Discovery}

		\author[1]{Faming~Xu\corref{cor1}}
		\author[1]{Yiding~Wang\corref{cor1}}
		\author[1]{Chen~Qiao\corref{cor2}}
		\ead{qiaochen@mail.xjtu.edu.cn}
		\author[2]{Gang~Qu}
		\author[3]{Vince~D.~Calhoun}
		\author[4]{Julia~M.~Stephen}
		\author[5]{Tony~W.~Wilson}
		\author[2]{Yu-Ping~Wang\corref{cor2}}
		\ead{wyp@tulane.edu}
		
		\cortext[cor1]{Equal Contribution}
		\cortext[cor2]{Corresponding author}
		\address[1]{School of Mathematics and Statistics, Xi'an Jiaotong University, Xi'an, 710049, China.}
		\address[2]{Department of Biomedical Engineering, Tulane University, New Orleans, LA 70118, USA.}
		\address[3]{Tri-Institutional Center for Translational Research in Neuroimaging and Data Science (TReNDS), Georgia State University, Georgia Institute of Technology, Emory University, Atlanta, GA 30030, USA.}
		\address[4]{Mind Research Network, Albuquerque, NM 87106, USA.}
		\address[5]{Institute for Human Neuroscience, Boys Town National Research Hospital, Boys Town, NE 68010, USA.}

		\begin{abstract}
			Dynamic effective connectivity networks (dECNs) reveal the changing directed brain activity and the dynamic causal influences among brain regions, which facilitate the identification of individual differences and enhance the understanding of human brain. Although the existing causal discovery methods have shown promising results in effective connectivity network analysis, they often overlook the dynamics of causality, in addition to the incorporation of spatio-temporal information in brain activity data. To address these issues, we propose a deep spatio-temporal fusion architecture, which employs a dynamic causal deep encoder to incorporate spatio-temporal information into dynamic causality modeling, and a dynamic causal deep decoder to verify the discovered causality. The effectiveness of the proposed method is first illustrated with simulated data. Then, experimental results from Philadelphia Neurodevelopmental Cohort (PNC) demonstrate the superiority of the proposed method in inferring dECNs, which reveal the dynamic evolution of directed flow between brain regions. The analysis shows the difference of dECNs between young adults and children. Specifically, the directed brain functional networks transit from fluctuating undifferentiated systems to more stable specialized networks as one grows. This observation provides further evidence on the modularization and adaptation of brain networks during development, leading to higher cognitive abilities observed in young adults.
		\end{abstract}
		
		
	\end{frontmatter}
	
	\section{Introduction}
	Brain is a highly complex system that consists of numerous brain regions and their connections. Dynamic effective connectivity network of brain regions characterizes the varying information flow and causal influences of brain \cite{zarghami2020dynamic}. Understanding causality is important for understanding and utilizing the intrinsic directed influence among brian activity data, making artificial intelligence techniques to process information like humans \cite{zhao2024coresets}. Causality learning includes both causal discovery and causal inference. Unlike causal inference, which aims to estimate outcome changes after certain variables are manipulated, causal discovery focuses on identifying the intrinsic causality based on observational data. In causal discovery, randomized controlled experiments are considered as the gold standard for discovering causality, but they are often expensive, time-consuming, unethical, or even impossible in many cases \cite{glymour2019review}. Thus, data-driven causal discovery have attracted significant attention in brain network analysis in recent years \cite{shojaie2022granger}.
	
	Functional magnetic resonance imaging (fMRI) is a non-invasive technique used to study the function and structure of the human brain \cite{2024The}. Its advantages include high spatial resolution and the ability to monitor brain activity in real time, allowing researchers to gather information among different brain regions. For time series data, time series causal discovery algorithms are used to infer the causality mainly based on Granger causal analysis. Tank et al. constructed a corresponding neural network for each time series and introduced sparse regularization to identify causality in nonlinear time series data \cite{tank2021neural}. Wu et al. identified the inherent causal relationship in nonlinear time series data by introducing disturbance information and minimizing mutual information loss \cite{wu2020discovering}. In brain connectivity analysis, Gadgil et al. proposed a model utilizing spatio-temporal graph convolution networks to analyze the functional connectivity from resting-state fMRI data \cite{gadgil2020spatio}. Alfakih et al. proposed a causal variational auto-encoder to study causal relationship between brain regions \cite{alfakih2023deep}. However, these methods aim to reveal fixed intrinsic relationship in data. On the other hand, dynamic functional connectivity is an interesting task in brain network analysis, which aims to discover the time-varying symmetric relationship among brain regions. Ji et al. used convolutional and recurrent neural networks to extract topology features to represent dynamic functional connectivity \cite{ji2024convolutional}. Xu et al. proposed to use graph convolutional network and convolutional neural networks to learn dynamic functional connectivity from multi-channel spatio-temporal data \cite{xu2024dynamic}.
	
	Although the aforementioned algorithms can infer the intrinsic causality or dynamic symmetric relationship in data, they still present the following issues:
	1) Current research focuses on fixed causality and does not consider causality changes over time. However, causality in real world are usually dynamic \cite{sachs2005causal}. Compared to fixed causality, dynamic causality further considers the evolution of causality over time, which better conforms to the real causality pattern. Therefore, a dynamic causality discovery method is required to explore how causality evolves regularly.
	2) There is a lack of research on causal discovery incorporating spatio-temporal information. Effective utilization of spatio-temporal information to understand spatio-temporal coupling relationships is crucial for discovering the causality in spatio-temporal data. However, current causal discovery algorithms predominantly emphasize utilizing spatial information or temporal information without fully taking advantage of spatio-temporal information.
	
	To address the above issues, a deep spatio-temporal fusion architecture, namely spatio-temporal dynamic causal deep auto-encoder (STDCDAE), is proposed. Because deep learning models have powerful feature extraction and representation capabilities and utilize nonlinear information effectively, STDCDAE adopts deep learning and is constructed based on a deep auto-encoder architecture to discover nonlinear causality accurately. Firstly, to discover the dynamic causality in data, the encoder with gated recurrent unit (GRU) is used, which generates dynamic causal mask matrices, and the decoder is used to verify the accuracy of the discovered causality. Subsequently, to discern causality within spatio-temporal data, we incorporate a spatio-temporal fusion strategy. Specifically, the encoder of STDCDAE contains a graph convolutional network (GCN) module together with GRU to fuse the spatio-temporal information of data to discover the causality, and the decoder contains a node graph convolutional network and a future forecasting multi-layer perceptron making use of the spatio-temporal information to verify the accuracy of the discovered causality. In addition, to utilize data structure information and prior knowledge to help discover causality accurately, data structure loss and causal divergence loss are introduced in the learning criteria of STDCDAE. In order to reduce the redundant noises and focus on the underlying causality, the sparse learning strategy is adopted in the learning process to promote the learning ability of the model.
	
	To validate the efficacy of the proposed model, STDCDAE is tested on linear and nonlinear simulated data; the experimental results show that our method can more accurately discover the causality in both the linear data and nonlinear data compared with other causal discovery algorithms. In addition, considering that the dynamic effective connectivity networks (dECNs) can reveal directed dynamic causality in brain functional networks, capture switching properties of the brain networks, and help understand the cognitive mechanisms of brain, the STDCDAE is applied to infer dECNs based on functional magnetic resonance imaging (fMRI) data from the Philadelphia Neurodevelopmental Cohort. Moreover, we aim to examine the alterations in brain effective connectivity throughout development by analyzing the spatial and temporal patterns of dECNs.
	
	\section{Methodology}
	Consider the sample set $X=\{x_1,x_2,\cdots,x_S\}$ with $S$ samples. For each sample $x_k\in \mathbb{R}^{N\times T \times d}$, let $N$ denote the number of nodes, $T$ denote the length of time series, and $D$ denote the feature dimension of each node at time $t$. For simplicity, we ignore the subscript $k$ and directly consider each sample $x$ in the below discussion. For sample $x$, $x^t_{i,j}$ represents the value of the $j$-th feature of the $i$-th node at time $t$, $x^t_i$ is used to represent the feature vector of node $i$ at time $t$, $x^t\in \mathbb{R}^{N\times d}$ is the feature matrix of sample $x$ at time $t$, and $x^{<t}\in \mathbb{R}^{N\times t-1\times d}$ is the feature tensor of sample $x$ from initial time to time $t-1$. The directed arrow in $x^{<t}\to x^t_i$ represents the directed influence relationship from historical cumulative information of all nodes to the $i$-th node at time $t$. If the potential relationship changes over time, it means there is a dynamic causality, otherwise, it means there is a fixed causality. To discover the dynamic causality among all nodes, the STDCDAE is proposed. 
	
	The STDCDAE comprises two parts: a dynamic causal deep encoder and a dynamic causal deep decoder. Specifically, to uncover the intrinsic causality in the data, the dynamic causal deep encoder which contains spatio-temporal information fusion module and mask matrix generation module is used to encode the data and obtain a causal mask matrix to infer the causality. Conversely, the dynamic causal deep decoder, furnished with a spatial information fusion module and a temporal information prediction module, which is used to forecast future information based on masked input information to construct evaluation indicators and evaluate the accuracy of causal mask matrices. The training for both encoder and decoder is supervised by the learning criteria consisting of reconstruction loss and causal constraint loss, which guide the model to discover causality in the data by utilizing data information and prior information. Compared with existing causal discovery algorithms, STDCDAE assimilates the congruence between historical cumulative spatio-temporal data and current representations by fusing the spatio-temporal information in order to discover dynamic causality in the spatio-temporal data. 
	
\begin{figure}
		\centering
		\includegraphics[width = 0.8\textwidth]{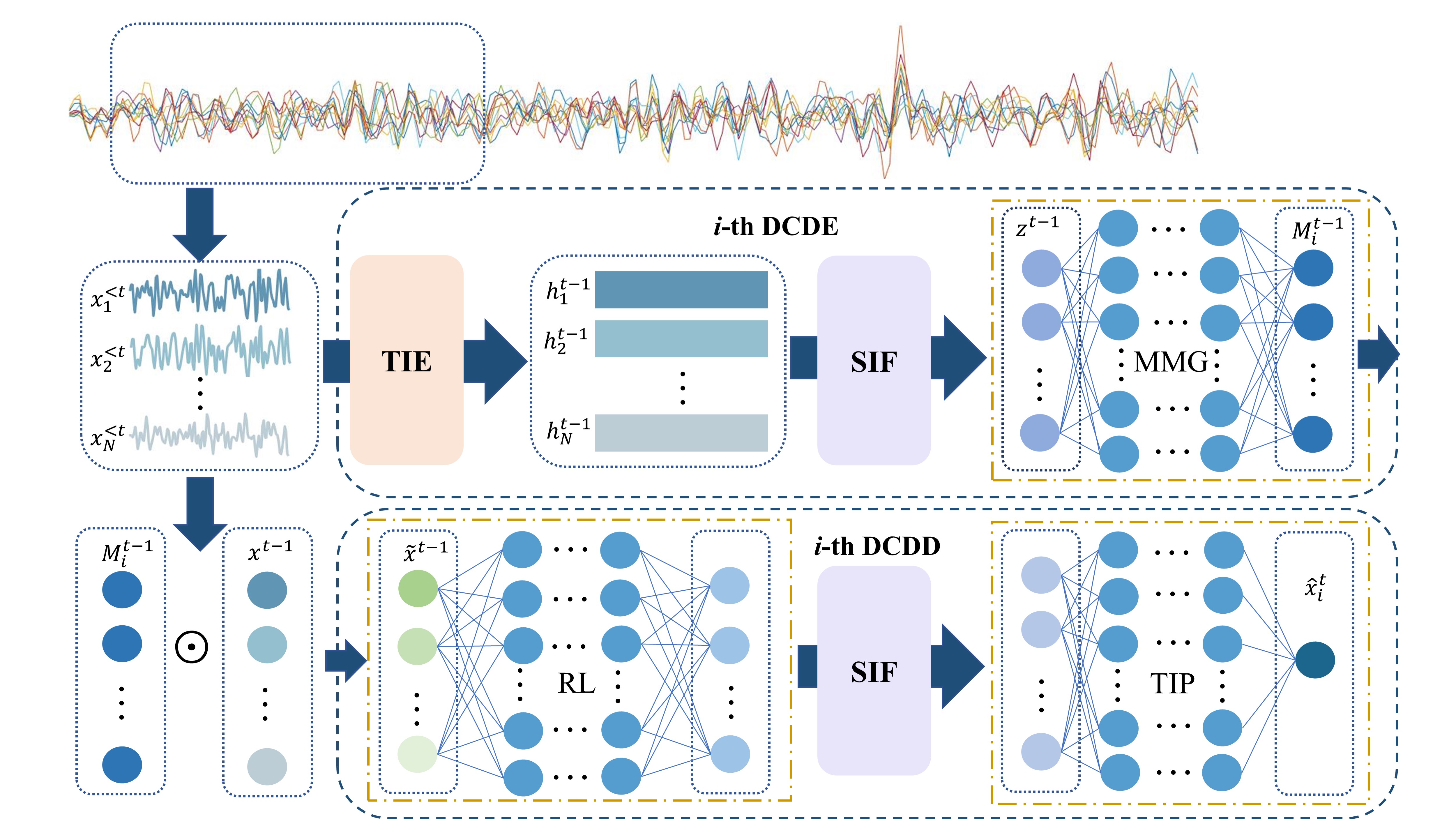}\\
		\includegraphics[width = 0.8\textwidth]{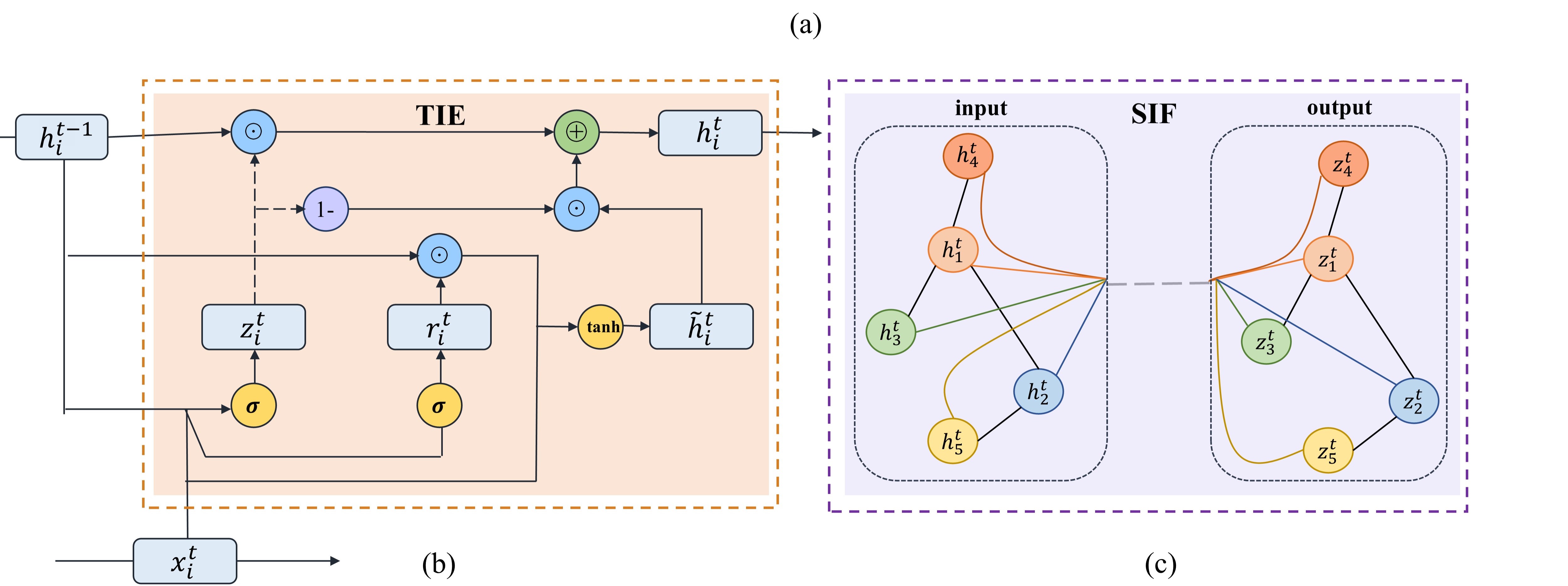}
		\caption{(a) The flow chart of STDCDAE, which demonstrates how the $i$-th deep auto-encoder in STDCDAE infers intrinsic dynamic causality for the $i$-th node. DCDE represents the dynamic causal deep encoder . DCDD represents the dynamic causal deep decoder. TIE and SIF represent temporal information extraction and spatial information fusion, respectively. MMG represents mask matrix generation. RL means representation learning. TIP represents temporal information processing. (b) The calculation process of the TIE module in STDCDAE. (c) The calculation process of the SIF module in STDCDAE.}
		\label{fig1}
	\end{figure}

	The proposed STDCDAE will be introduced in the following aspects. Firstly, the SIF and TIE modules used in STDCDAE are introduced. Then, we describe the details of the dynamic causal deep encoder and dynamic causal deep decoder. Finally, the learning criteria of STDCDAE are discussed. The flow chart of STDCDAE is shown in Fig. \ref{fig1}.
	\subsection{The Architecture of SIF and TIE}
	\subsubsection{\textbf{Spatial Information Fusion (SIF)}}
	The aim of spatial information fusion module is to integrate information from nodes under a target graph with adjacency matrix. Convolutional neural networks are usually used in many fields, however, compared to the conventional convolutional neural networks, which only handle the data in Euclidean space, GCNs have attracted significant attention due to their robust ability to process non-Euclidean graph data \cite{bai2020learning}. For the graph $G=\{V, E\}$, where $V=\{V_1,\cdots, V_N\}$ represents the set of vertices in the graph and $E=\{E_1,\cdots,E_N\}$ represents the set of edges. The feature vector for the $i$-th node in $V$ is defined as $H_i\in\mathbb{R}^{d}$ and the feature matrix is represented as $H=(H_1,\cdots,H_N)\in\mathbb{R}^{N\times d}$. In addition, the weighted adjacency matrix between nodes is defined as $A\in \mathbb{R}^{N\times N}$, where $A_{ij}$ represents the connection between node $i$ and node $j$. The general propagation rule \cite{zhang2023explainable} of GCN is
	\begin{equation}
		\label{eq1}
		H^{(l+1)}=\phi(\tilde{D}^{-\frac12}\tilde{A}\tilde{D}^{-\frac12}H^{(l)}W^{(l)})
	\end{equation}
	Here $H^{(l)}$ and $H^{(l+1)}$ represent the features of the $l$-th layer and the $l+1$-th layer, respectively. $W^{(l)}$ denotes the weight between the $l$-th layer and the $l+1$-th layer. The $\tilde{A}=A+\lambda I_N$ is the adjacency matrix affected by self-loop, where $I_N$ is a $N$-dimensional identity matrix, and $\lambda$ characterizes the self-loop intensity. The degree matrix $\tilde{D}$ diagonal, with its diagonal element given by $\tilde{D}_i=\sum_{j}\tilde{A}_{ij}$. $\phi(\cdot)$ is a nonlinear activation function.
	
	\subsubsection{\textbf{Temporal Information Embedding (TIE)}}
	The aim of temporal information embedding module is to extract low dimensional representation from time series. Recurrent neural networks are commonly used in time series analysis, however, compared with traditional recurrent neural network, GRU introduces a gate mechanism to avoid long-term dependency issues while memorizing the sequence information, and is widely used in time series analysis \cite{ahmad2022novel}. Compared with long short term memory network (LSTM), GRU combines the forget gate and input gate into an update gate. In this case, GRU has a lower risk of overfitting and is computationally more efficient than LSTM due to having fewer parameters. The update formula of GRU is
	\begin{equation}
		h_t=z_t\odot h_{t-1}+(1-z_t)\odot \tilde{h}_t
	\end{equation}
	where $h_t$, $h_{t-1}$ represent the current and previous memory cell states. $z_t$, $\tilde{h}_t$ represent the values of the update gate and candidate memory cells at current time and are calculated by
	\begin{equation}
		\begin{aligned}
			&z_t=\sigma(W_zx_t+U_zh_{t-1}+b_z)\\
			&\tilde{h}_t=\text{tanh}(W_hx_t+U_h(r_t\odot h_{t-1})+b_h)
		\end{aligned}
	\end{equation}
	in which $W_z$, $U_z$, $W_h$, $U_h$ and $b_z$, $b_h$ are the weights and biases for update gate and candidate memory cell, respectively. $x_t$ represents the input information at the current time, and $r_t$ represents the reset gate state calculated by
	\begin{equation}
		r_t=\sigma(W_rx_t+U_rh_{t-1}+b_r)
	\end{equation}
	where, $W_r$, $U_r$ and $b_r$ represent the weights and bias of the reset gate. $x_t$ represents the input information at the current time. $h_{t-1}$ represents the state of the memory cell at the previous time.
	
	\subsection{STDCDAE}
	The STDCDAE consists of two parts: the dynamic causal deep encoder which fuses the spatio-temporal information and discovers the dynamic causality, and the dynamic causal deep decoder which verifies the accuracy of the discovered causality. For each node $i\quad(i=1,\cdots,N)$ the dynamic causal deep encoder $f_{en,i}(\cdot)$ and the dynamic causal deep decoder $f_{de,i}(\cdot)$ is applied to discover the intrinsic causality between the $i$-th node and the other nodes, so that the effects from inputs to the $i$-th node and other nodes can be easily disentangled. This makes the causality between different nodes can be inferred clearly and accurately.
	
	Next, the dynamic causal deep encoder, the dynamic causal deep decoder, and the learning criteria of STDCDAE will be introduced.
	
	\subsubsection{\textbf{Dynamic Causal Deep Encoder (DCDE)}}
	For effectively utilizing the spatio-temporal information in the observed data and discovering the dynamic causality between nodes in the data, the dynamic causal deep encoder is constructed. 
	
	For the $i$-th node $x_i$ of sample $x$, its corresponding dynamic causal deep encoder $f_{en,i}(\cdot)$ is established to discover the causality between $x^{t-1}$ and $x^t_i$ through utilizing the spatio-temporal information in the observed data. Meanwhile, in order to utilize comprehensive historical information to help accurately discover the current causality, the preceding time series are used as the inputs of DCDE to generate current causal mask. The causality is inferred based on mask vector $M^{t-1}_{i}\in \mathbb{R}^{N\times 1}$. The $j$-th element $M^{t-1}_{ij}$ of $M^{t-1}_i$ characterizes whether there exists causality between $x^{t-1}_j$ and $x^t_i$. If $M^{t-1}_{ij}\ne 0$, then $x^{t-1}_j$ is regarded as cause of $x^t_i$, $M^{t-1}_{ij}=0$ indicates there is no intrinsic causality between $x^{t-1}_j$ and $x^t_i$. The DCDE contains a spatio-temporal information fusion module and a mask matrix generation module.
	
	\textit{\textbf{i) Spatio-temporal information fusion (STIF):}} In order to fuse the spatio-temporal information to obtain better representations for causal discovery, the spatio-temporal fusion strategy is adopted. For the $j$-th node, GRU is applied to learn the current representation from the historical information.
	\begin{equation}
		\label{eq2}
		h^{t-1}_j=\text{GRU}_j(x^{<t}_j)
	\end{equation} 
	The current representation vector $h^{t-1}_j\in \mathbb{R}^{1\times d_1},\quad(j=1,2,\cdots,N)$ is concatenated in rows to obtain the time information matrix $h^{t-1}=(h^{t-1};\cdots;h^{t-1}_N)\in \mathbb{R}^{N\times d_1}$. The spatial information in the $h^{t-1}$ is fused through GCN. 
	\begin{equation}
		\label{eq3}
		z^{t-1}=\phi(\tilde{L}h^{t-1}W)
	\end{equation} 
	where $\tilde{L}=\tilde{D}^{-\frac{1}{2}}\tilde{A}\tilde{D}^{-\frac{1}{2}}$ is the graph convolution kernel, and $\tilde{D}$ and $\tilde{A}$ are defined in Eq. (\ref{eq1}). The $z^{t-1}\in\mathbb{R}^{N\times d_2}$ is regarded as a representation, which not only considers current and historical information but also integrates information from domain nodes. This allows it to better characterize the evolution of causality between nodes over time.
	
	\textit{\textbf{ii) Mask matrix generation (MMG)}}: In order to reveal the dynamic causality between nodes, the MLP $g_{en,ge}(\cdot)$ is applied to generate the $i$-th row $M^{t-1}_i$ in the mask matrix $M^{t-1}$, and forecast whether there exists causality between the $i$-th node and other nodes.
	\begin{equation}
		\label{eq4}
		M^{t-1}_i=\sigma(g_{en,ge}(z^{t-1},\theta_{en,ge}))
	\end{equation}
	where, $\sigma(\cdot)$ is the sigmoid function, with $\theta_{en,ge}$ being the parameters of the MLP. By integrating the historical and spatial information of each node at all times, the causal mask vector $M^{t-1}_i$ is obtained to infer the causality between the $i$-th node and other nodes.  
	
	Based on Eq. (\ref{eq2})-(\ref{eq4}), the dynamic causal deep encoder $f_{en,i}(\cdot)$ is constructed, and the causal mask vector is generated through
	\begin{equation}
		M^{t-1}_i = f_{en,i}(x^{<t})
	\end{equation}
	According to the generated causal mask vector $M^{t-1}_i$, the following mask operations are performed
	\begin{equation}
		\label{eq6}
		\tilde{x}^{t-1}=[M^{t-1}_{i1}\odot x^{t-1}_1,\cdots,M^{t-1}_{iN}\odot x^{t-1}_N]
	\end{equation}
	$\tilde{x}^{t-1}$ will be applied as input to the next dynamic causal deep decoder to verify the accuracy of the causal mask vector.
	
	\subsubsection{\textbf{Dynamic Causal Deep Decoder (DCDD)}}
	If there is no causality between $x^t_i$ and $x^{t-1}_j (t=1,\cdots,T)$, the mask-out of $x^{t-1}_j$ does not affect the prediction of $x^t_i$. To verify whether the causality between $x^t_i$ and $x^{t-1}_j (j=1,\cdots,N)$ has been captured by the causal mask vector $M^{t-1}_i$, the dynamic causal deep decoder $f_{de,i}(\cdot)$ is constructed. 
	
	Considering the causality between nodes in observed data is usually unknown, it is not possible to determine whether the accurate causality is captured through data. Therefore, the similarity between the true value $x^t_i$ and the output, which is calculated by taking the $\tilde{x}^{t-1}$ obtained from Eq. (\ref{eq6}) as input to dynamic causal deep decoder $f_{de,i}(\cdot)$, is used to evaluate the accuracy of the inferred causality. Meanwhile, considering the temporal and spatial characteristics of the data, the dynamic causal deep decoder is constructed based on spatio-temporal mechanism modeling. The DCDD contains a spatial information fusion module and a temporal information prediction module.
	
	\textit{\textbf{i) Spatial information fusion (SIF):}} In order to obtain a better representation for prediction, spatial information fusion is performed. Before the spatial information fusion, a MLP $g_{de,re}(\cdot)$ is applied to perform representation learning on the inputs to obtain more suitable representations for spatial information fusion. 
	\begin{equation}
		\label{eq7}
		h^{t-1}_j=g_{de,re}(\tilde{x}^{t-1}_j,\theta_{de,re})
	\end{equation} 
	where $\theta_{de,re}$ is the parameters of the MLP. After representation learning, $h^{t-1}_{j}\in\mathbb{R}^{1\times d_3}$ is concatenated as the feature matrix $h^{t-1}=[h^{t-1}_1,\cdots,h^{t-1}_N]\in\mathbb{R}^{N\times d_3}$, which is the input for the NGCN to fuse the spatial information.
	\begin{equation}
		z^{t-1}=\phi(\tilde{L}_ih^{t-1}W_n)
	\end{equation}
	where $\phi(\cdot)$ is the nonlinear activation function, $\tilde{L}_i$ is the $i$-th row of the graph convolution matrix $\tilde{L}$ which is shown in Eq. (\ref{eq2}), and $W_n$ is the weight of the NGCN. $z^{t-1}$ is regarded as a representation aggregates the spatial information.
	
	\textit{\textbf{ii) Temporal information prediction (TIP):}} To verify the accuracy of the causal mask matrix, the temporal information prediction network is applied to forecast the next time information of the $i$-th node based on the current representation $z^{t-1}\in \mathbb{R}^{1\times d_4}$ which aggregates the domain nodes information.
	\begin{equation}
		\label{eq9}
		\hat{x}^t_i=g_{de,pr}(z^{t-1},\theta_{de,pr})
	\end{equation}
	where $g_{de,pr}(\cdot)$ is the temporal information prediction network which implement a MLP, and $\theta_{de,pr}$ is its parameters. $\hat{x}^t_i$ is the predictive value of next time information of the $i$-th node.
	
	Based on Eq. (\ref{eq7})-(\ref{eq9}), the dynamic causal deep decoder $f_{de,i}(\cdot)$ is constructed, and the predictive value of next time information of the node is calculated by
	\begin{equation}
		\hat{x}^t_i=f_{de,i}(\tilde{x}^{t-1})
	\end{equation}
	The estimated $\hat{x}^t_i$ is applied to construct the learning criteria of STDCDAE to guide model training.
	
	\subsubsection{\textbf{The Learning Criteria of STDCDAE}}
	The learning criteria of STDCDAE for all nodes in the sample set are
	\begin{equation}
		\begin{aligned}
			\mathcal{L} =\sum^{N}_{i=1}\left(\mathcal{L}_{\text{recon},i}+\beta_1\mathcal{L}_{\text{struct},i}+\beta_2\mathcal{L}_{\text{divergence},i}+\beta_3\mathcal{L}_{\text{sparsity},i}\right)
		\end{aligned}
	\end{equation}
	The loss function $\mathcal{L}$ is calculated by summing the losses of all nodes. For the $i$-th node $x_i$, $\mathcal{L}_{\text{recon},i}$ is the reconstruction loss and $\mathcal{L}_{\text{struct},i}$ is the structure loss, both of which are the losses for the reconstruction data. $\mathcal{L}_{\text{divergence},i}$ is the distribution divergence loss, and $\mathcal{L}_{\text{sparsity},i}$ the causal sparsity loss, both of which are the constraint losses for the causal mask vectors. The $\beta_1,\beta_2,\beta_3$ are the coefficients of these losses.
	
	The reconstruction loss $\mathcal{L}_{\text{recon},i}$ is applied to guide the model in discovering the intrinsic causality of data. Because the accuracy of causality inferred by STDCDAE is evaluated by the error between the output $\hat{x}^t_i$ of $f_{de,i}(\tilde{x}^{t-1})$ and $x^t_i$, the reconstruction function is
	\begin{equation}
		\mathcal{L}_{\text{recon},i}=\frac{1}{T-1}\sum^T_{t=2}||x^t_i-\hat{x}^t_i||^2_2
	\end{equation}
	
	Considering that combining the structure information of the data can help the model more effectively discover causality in the data \cite{Goudet2018}, the following structure loss is proposed
	\begin{equation}
		\mathcal{L}_{\text{struct},i}=\frac{1}{N}\frac{1}{T-1}\sum^T_{t=2}\sum^N_{j=1}(\tau(x^t_j,x^t_i)-\tau(x^t_j,\hat{x}^t_i))^2
	\end{equation}
	The structure loss considers the structure of data at each moment, which helps the model discover the causality more accurately and effectively. $\tau(x,y)=\exp(-\gamma||x-y||^2_2)$ with $\gamma$ being the scale parameter.
	
	To consider the prior knowledge of causality distribution and improve the model accurately discover the causality in the data, the following causal distribution divergence loss are introduced
	\begin{equation}
		\mathcal{L}_{{div},i} = \lambda_1\mathcal{L}_{{E-div,i}} + \lambda_2\mathcal{L}_{{KL-div,i}}+
		\lambda_3\mathcal{L}_{{JS-div,i}}
	\end{equation}
	where the $\mathcal{L}_{{E-div,i}}$ is the entropy, which is used to reduce the uncertainty of the causal mask vector when prior knowledge of causality is unknown. The $\mathcal{L}_{{KL-div}}$ and $\mathcal{L}_{{JS-div,i}}$ are Kullback-Leibler (KL) divergence and Jensen-Shannon (JS) divergence, respectively. Both of them are used to measure the differences between the distribution of causal mask vector and the causality prior distribution. The $\lambda_1,\lambda_2,\lambda_3$ is the weight coefficients of different distribution divergence, and $\lambda_1 +\lambda_2+\lambda_3=1$. $\mathcal{L}_{{E-div,i}}$, $\mathcal{L}_{{KL-div,i}}$, and $\mathcal{L}_{{JS-div,i}}$ are calculated as
	\begin{equation}
		\begin{aligned}
			&\mathcal{L}_{{E-div,i}}=-\frac{1}{N}\sum^N_{j=1}\tilde{M}_{ij}\log\tilde{M}_{ij}\\
			&\mathcal{L}_{{KL-div,i}}=\frac{1}{N}\sum^N_{j=1}\tilde{M}_{ij}\log\frac{\tilde{M}_{ij}}{P_{ij}}\\
			&\mathcal{L}_{{JS-div,i}}=\frac{1}{2N}\sum^N_{j=1}(\tilde{M}_{ij}\log\frac{\tilde{M}_{ij}}{Q_{ij}}+P_{ij}\log\frac{P_{ij}}{Q_{ij}})
		\end{aligned}
	\end{equation}
	where $\log\tilde{M}_{ij}=\frac{1}{T-1}\sum^T_{t=2}M^{t-1}_{ij}$, $P_{ij}$ is the prior distribution of $M_{ij}$, and $Q_{ij}=\frac{P_{ij}+\tilde{M}_{ij}}{2}$.
	
	Because the sparse learning strategy allows the model to ignore the redundant and unimportant features and focus on the basic features, which is helpful for improving ability of causal discovery and reducing uncertainty in causal discovery, the causal sparsity loss $\mathcal{L}_{\text{sparsity},i}$ is introduced. Considering that compared to traditional $L_1$ regularization and $L_2$ regularization, log-sum regularization has a better approximation ability to $L_0$ regularization and stronger sparse learning ability, which is suitable for the causal learning among high dimensional data \cite{qiao2020log}, the log-sum regularization is applied as the causal sparsity constraint for $M^{t-1}_{ij}$. The causal sparsity loss is
	\begin{equation}
		\mathcal{L}_{\text{sparsity},i}=\frac{1}{N}\frac{1}{T-1}\sum^N_{j=1}\sum^T_{t=2}\log(\frac{|M^{t-1}_{ij}|}{\epsilon}+1)
	\end{equation}
	where $\epsilon$ is the scale parameter of log-sum regularization. By introducing the causal sparsity constraint, $M^{t-1}_{ij}$ is close to 1 when $x^{t-1}_j$ is the cause of $x^t_i$ and $M^{t-1}_{ij}$ is close to 0 when $x^{t-1}_j$ is not the cause of $x^t_i$.
	
	The STDCDAE is trained by back-propagation algorithms based on the above learning criteria to extract spatio-temporal information in data, and infer the causal mask matrices $M^t = [M^t_1,\cdots,M^t_N]\in\mathbb{R}^{N\times N} (t=1,\cdots,T),$ which represent the dynamic causality in the observed data.
	\section{Simulation Experiments}
	In order to verify the performance of the proposed STDCDAE,  we first applied it to two simulated datasets, i.e., the linear vector autoregression (VAR) data \cite{tank2021neural}, and the nonlinear Lorenz-96 data \cite{karimi2010extensive}, to discover the causality in the data. The results show that the proposed model can accurately discover the causality in both linear data and nonlinear data.
	
	\subsection{VAR Experiments}
	To assess the performance of the proposed STDCDAE on linear data, we applied it to the VAR-1 and VAR-2 datasets from \cite{tank2021neural}, each consisting of 20 variables. These datasets were simulated using randomly generated sparse transition matrices. The VAR-1 is the VAR with a lag order of 1, and the VAR-2 is the VAR with a lag order of 2. The STDCDAE is compared with the cMLP model and cLSTM model \cite{tank2021neural}, the LOO-LSTM model, and IMV-LSTM model \cite{guo2019exploring}. To maintain evaluation consistency, the area under the receiver operating characteristic curve (AUROC) was employed as the evaluation metric. The presented results are the averages from five experiments, each conducted with a different randomly set seed.  The experiment results are shown in Table \ref{table1}.
	
	\begin{table*}[t]
		\caption{Comparison of AUROC for identifying causality in linear data among different models in different VAR lag orders and length of the time series (mean $\pm$ 95\% confidence intervals)\label{table1}}
		\centering	
		\begin{threeparttable}
			\footnotesize
			\begin{tabular}{l c c c c c c}
				\hline
				Model& \multicolumn{3}{c}{VAR-1} &  \multicolumn{3}{c}{VAR-2}\\ 
				\cmidrule(r){1-1} \cmidrule(r){2-4}	\cmidrule(r){2-4} \cmidrule(r){5-7}
				$T$&250&500&1000&250&500&1000\\
				\hline
				STDCDAE&\textbf{96.7$\pm$0.2}&\textbf{99.0$\pm$0.1}&\textbf{99.9$\pm$0.1}&\textbf{86.5$\pm$0.2}&\textbf{94.1$\pm$0.1}&\textbf{98.9$\pm$0.1}\\
				cMLP&91.6$\pm$0.4&94.9$\pm$0.2&98.4$\pm$0.1&84.4$\pm$0.2&88.3$\pm$0.4&95.1$\pm$0.2\\
				cLSTM&88.5$\pm$0.9&93.4$\pm$1.9&97.6$\pm$0.4&83.5$\pm$0.3&92.5$\pm$0.4&97.8$\pm$0.1\\
				IMV-LSTM&53.7$\pm$7.9&63.2$\pm$8.0&60.4$\pm$8.3&53.5$\pm$3.9&54.3$\pm$3.6&55.0$\pm$3.4\\
				LOO-LSTM&50.1$\pm$2.7&50.2$\pm$2.6&50.5$\pm$1.9&50.1$\pm$1.4&50.4$\pm$1.4&50.0$\pm$1.0\\
				\hline		
			\end{tabular}
			\begin{tablenotes}
				\item[1] The results of cMLP, cLSTM, IMV-LSTM, LOO-LSTM here are sourced from \cite{tank2021neural}.
				\item[2] $T$ denotes the length of the time series. 
			\end{tablenotes}
		\end{threeparttable}
	\end{table*}
	
	For STDCDAE, the number of hidden units is set to be 15 for the dynamic causal deep encoder and the dynamic causal deep decoder for each node. The Adam algorithm is applied for the network parameter updates. Since the grid search method can make a comprehensive search over a given hyperparameters space and be parallelized to find more stable optimal hyperparameters, it is used to select hyperparameters \cite{saud2020performance}. Through the grid search method, the learning rate of Adam algorithm is set as $1\times 10^{-3}$, the coefficients of structural loss, distribution divergence loss and causal sparsity loss $\beta_1,\beta_2,\beta_3$ are set as 0.01, 0.35, and 0.35, respectively.
	
	The experimental results in Table \ref{table1} demonstrate that compared with other methods, the proposed STDCDAE can more accurately discover the causality in the linear data under different time series lengths and lag orders. The 95\% confidence intervals of STDCDAE are relatively narrower than other methods, which means that the proposed method is more robust than other methods. In addition, the ability of STDCDAE to identify causality rapidly improves, with the length of time series increasing. The complicated causality of the VAR-2 data can also be accurately revealed by the proposed STDCDAE in sufficient length of time series.
	\subsection{Lorenz-96 Experiments}
	The proposed STDCDAE is applied in the nonlinear Lorenz-96 data with $N$ variables to verify the performance of the proposed STDCDAE in nonlinear data. The Lorenz-96 data is generated by
	\begin{equation}
		\frac{\text{d}x_{ni}}{\text{d}t}=(x_{t(n+1)}-x_{t(n-2)})x_{t(n-1)}-x_n+F
	\end{equation}
	where $x_{t(-1)}=x_{t(N-1)},x_{t0}=x_{tN},x_{t(N+1)}=x_{t1}$, and $F$ is a parameter that determines the degree of nonlinearity and chaos in the time series. In the Lorenz-96 experiments, in addition to verifying the ability of STDCDAE to discover causality under different time series lengths, the impact of data complexity, i.e., the number of variables on the accuracy of revealing the causality in the data is also considered. In this experiment, the parameter $F$ is set to be 10, and the number of variables $N\in\{10,20,30\}$, and the length of time series $T\in\{250,500,1000\}$. The cMLP model, cLSTM model, the VAR-LiNGAM model proposed in \cite{hyvarinen2010estimation} and the VAR model are applied as the benchmark methods. Consistent with the VAR experiments, AUROC is used as the evaluation indicator in the Lorenz-96 experiments. The results are the means of five repeated experiments with different random seeds which are randomly set. Experimental results are shown in Table \ref{table2}.
	\begin{table*}[h]
		\caption{Comparison of AUROC for identifying causality in nonlinear data among different models in different the number of variables and length of the time series (mean $\pm$ 95\% confidence intervals)\label{table2}}
		\centering	
		\begin{threeparttable}
			\footnotesize
			\begin{tabular}{lcccc}
				\hline
				Model&$T$&250&500&1000\\
				\hline
				STDCDAE& \multirow{5}{*}{$n=10$}&\textbf{98.59$\pm$0.14}&\textbf{99.84$\pm$0.02}&\textbf{99.91$\pm$0.01}\\
				cLSTM& &89.19$\pm$0.60&89.82$\pm$0.48&89.69$\pm$0.19\\
				cMLP& &80.41$\pm$0.53&81.20$\pm$0.33&80.96$\pm$0.15\\
				VAR-LiNGAM& &66.97$\pm$0.17&68.53$\pm$0.19&69.65$\pm$0.10\\
				VAR& &68.49$\pm$0.45&69.71$\pm$0.29&71.31$\pm$0.19\\
				\hline
				STDCDAE& \multirow{5}{*}{$n=20$}&\textbf{94.41$\pm$0.17}&\textbf{99.07$\pm$0.06}&\textbf{99.94$\pm$0.01}\\
				cLSTM& &87.42$\pm$0.37&90.07$\pm$0.17&91.45$\pm$0.13\\
				cMLP& &80.25$\pm$0.20&81.77$\pm$0.21&83.23$\pm$0.09\\
				VAR-LiNGAM& &69.11$\pm$0.18&71.30$\pm$0.08&72.95$\pm$0.05\\
				VAR& &66.60$\pm$0.21&70.38$\pm$0.12&73.02$\pm$0.10\\
				\hline
				STDCDAE& \multirow{5}{*}{$n=30$}&\textbf{87.23$\pm$0.21}&\textbf{98.82$\pm$0.04}&\textbf{99.93$\pm$0.01}\\
				cLSTM& &85.11$\pm$0.17&89.87$\pm$0.11&91.16$\pm$0.07\\
				cMLP& &79.34$\pm$0.20&81.17$\pm$0.10&80.90$\pm$0.09\\
				VAR-LiNGAM& &69.29$\pm$0.08&71.54$\pm$0.06&73.45$\pm$0.02\\
				VAR& &68.23$\pm$0.24&70.77$\pm$0.05&72.24$\pm$0.04\\
				\hline
			\end{tabular}
		\end{threeparttable}
	\end{table*}
	
	For STDCDAE, we set the number of hidden units to 20 for both the dynamic causal deep encoder and the dynamic causal deep decoder for each node. The Adam algorithm is applied for the network parameter updating. Through the grid search method, the learning rate of Adam algorithm is set as $1\times 10^{-3}$, the coefficients of structural loss, distribution divergence loss and causal sparsity loss $\beta_1,\beta_2,\beta_3$ are set to 10, 0.5, and 0.5.
	
	Results from the Lorenz-96 experiments in Table \ref{table2} demonstrate that STDCDAE outperforms other benchmark methods in accurately uncovering intrinsic causality across varying numbers of nodes and time series lengths. Specifically, AUROC of STDCDAE is superior to other benchmark methods for different nodes and time series length. In addition, the 95\% confidence interval of STDCDAE is narrower than that of other benchmark methods, underscoring its robustness for nonlinear data. At the same time, when the length of time series is sufficient, our method can accurately reveal the causality in the observed data. Notably, the ability of STDCDAE to identify causality in the data rapidly improves with the increase of temporal length.
	
	\section{Analysis of Dynamic Effective Connectivity Network in Brain}
	Compared with the functional connectivity network (FCN) based on the correlation between regions of interest (ROIs), the effective connectivity network (ECN) based on the causality between ROIs considers not only the connectivity strength but also the information flow between different ROIs \cite{friston2011functional}. The dynamic ECN (dECN), which is obtained based on the causality between ROIs at different times, further reflects the transformation of ECN over time. This captures the switching properties of the brain directed networks and reveals the variation of causality relationship among different brain networks, which is a new perspective for the research on brain functional networks.
	
	In this section, the proposed STDCDAE is applied to the resting state fMRI (rs-fMRI) data to infer the dECNs of children and young adults. Our aim is to investigate age-related changes in the dECNs and to discern variances in the information interaction patterns of brain functional networks between different age groups. Firstly, data preprocessing and hyperparameter selection are introduced. Then, the dECNs inferred by STDCDAE are compared with the ECNs and FCNs inferred by other methods to verify the performance of the STDCDAE. Next, the differences in interaction patterns of the brain functional networks between different groups are analyzed from the representation of dECNs and time-varying patterns of dECN, in order to reveal the changes of brain functional networks with brain development. 
	
	\subsection{Data Preprocessing and Model Hyperparameters}
	This research is based on the Philadelphia Neurodevelopmental Cohort (PNC) data. PNC is a large scale collaborative project between the Brain Behavior Laboratory at the University of Pennsylvania and the Children's Hospital of Philadelphia, which contains rs-fMRI data from 193 children aged 103-144 months and 204 young adults aged 216-271 months \cite{Zille2018FusedEO, wang2024deep}. For the collected data, first, the statistical parametric mapping 12 is applied for implementing the standard brain imaging preprocessing, which includes the motion correction, spatial normalization, and spatial smoothing with a 3mm full width half maximum Gaussian kernel \cite{chen2024explainable}. Next, a regression procedure is employed to remove the influence of motion, and a band-pass filter is applied to the functional time series within a frequency range of 0.01 Hz to 0.1 Hz. Finally, according to the definition of brain regions by \cite{power2011functional}, the standard 264 ROIs were used for the dimensionality reduction of the data. For each subject, a $264\times T$ matrix is obtained, where 264 represents the number of ROIs, and $T=124$ represents the number of time points.
	
	For the preprocessed data, the proposed STDCDAE is applied to infer the brain ECN of children and young adults separately. For STDCDAE, the number of hidden units is set to 50 for the dynamic causal deep encoder and the dynamic causal deep decoder of all nodes. The Adam algorithm is applied for the network parameter updates. Through the grid search method, the learning rate of the Adam algorithm is set to $1\times 10^{-3}$, the coefficients of structural loss, distribution divergence loss and causal sparsity loss $\beta_1,\beta_2,\beta_3$ are set to $1\times 10^{-3}$, $1\times 10^{-2}$, and $5\times 10^{-3}$, respectively. 
	
	\subsection{Inference of Brain dECN}
\begin{figure*}
		
		\includegraphics[width=\linewidth]{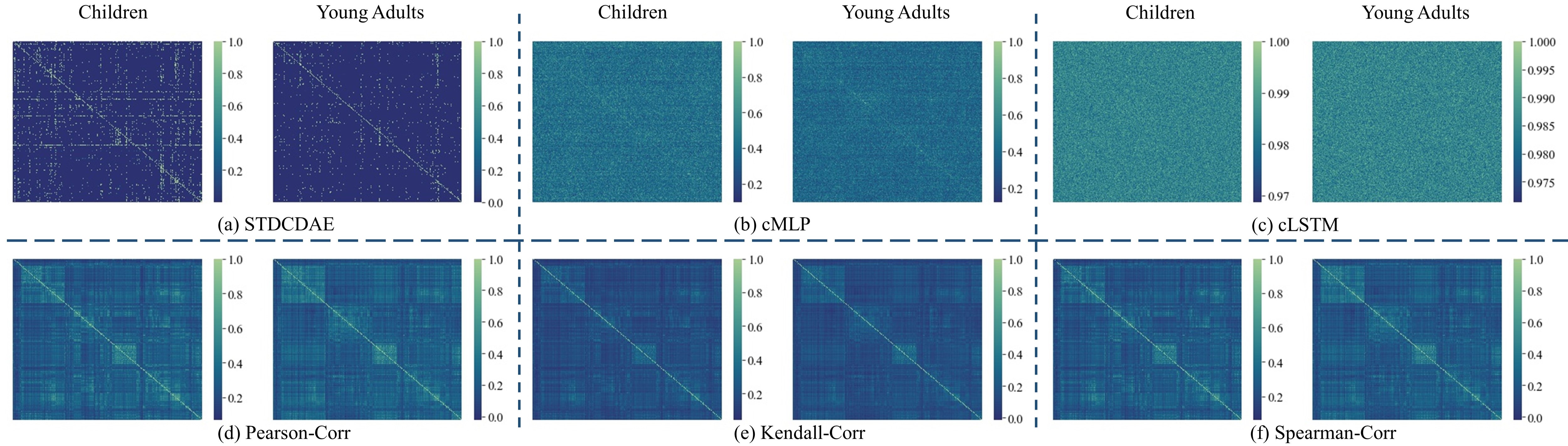}
		\caption{The brain functional networks of children and young adults inferred by different methods, where the first row shows the ECNs inferred by different causal discovery algorithms, and the second row shows the FCNs inferred by different correlation coefficients. Compared with cMLP and cLSTM, STDCDAE reveals clearer different ECN patterns between groups. Meanwhile, compared with the FCNs revealed by correlation coefficients, the ECNs inferred by STDCDAE not only capture the directed relationship among ROIs, but also exhibit higher sparsity, which is helpful for revealing the essential differences between different groups.}
		\label{fig2}
	\end{figure*}

	The mask matrix inferred by STDCDAE in the time $t$ is defined as the dECN at the time $t$, which characterizes the dynamic causality between ROIs. To verify the performance of the proposed STDCDAE in the inference of brain functional network. The ECNs inferred by STDCDAE, which are obtained by averaging the dECNs in the time dimension, are compared with the ECNs inferred by the cMLP model and cLSTM model, and the FCNs inferred by Pearson correlation coefficient, Kendall correlation coefficient, and Spearman correlation coefficient. The brain functional network patterns inferred by the above methods are shown in Fig. \ref{fig2}.
	
	The experimental results shown in Fig. \ref{fig2} indicate that the proposed STDCDAE can better capture the connectivity patterns between different ROIs compared to cMLP and cLSTM. Fig. \ref{fig2} (b) and (c) show that the brain functional network inferred by cMLP and cLSTM can not capture the connectivity patterns between ROIs well. This may be led by the high-dimensional but small-sample-size property of rs-fMRI, which makes cMLP and cLSTM difficult to accurately capture the intrinsic causality in the rs-fMRI data based solely on single-layer network weights and sparse constraints. The comparison results with FCNs inferred by different correlation coefficients show that the ECNs inferred by STDCDAE not only capture the connectivity patterns between ROIs well, but also have good sparsity which is helpful for effectively identifying the differences in brain functional networks between children and young adults. The FCNs inferred by the Pearson correlation coefficient, Kendall correlation coefficient, and Spearman correlation coefficient can characterize the connectivity patterns between different ROIs, but it is not as effective as the STDCDAE in characterizing the differences in brain functional networks between different groups and sparsity. In addition, the brain functional networks inferred by the correlation coefficient only consider that the correlation between different ROIs, resulting undirected networks which makes it not suitable for analyzing the information flow between different ROIs. The brain functional networks inferred by the proposed STDCDAE are based on the causal relationship between ROIs, which can effectively reveal the  information interaction patterns between ROIs based on the causality, thus contributing to a deeper exploration of the mechanisms of brain function.
	
	In order to further demonstrate that the proposed STDCDAE can better characterize the differences in brain functional networks between children and young adults, the  Structural Similarity (SSIM), Cosine Similarity (CS), Jenson Shannon Divergence (JSD), and Structural Euclidean Distance (SED) are chosen to measure the differences in ECNs. Meanwhile, the accuracy of classification (ACC) based on the support vector machines is also used to evaluate the differences in ECNs between different groups in Table \ref{table4}.
	
	\begin{table*}
		\caption{Differences in functional networks inferred by different methods between children and young adults}
		\label{table4} \centering
		\begin{center}
			\footnotesize
			\begin{tabular}{lccccc}
				\hline 
				Method&SSIM&CS&JSD&SED&ACC\\
				\hline
				STDCDAE&\textbf{0.114}&\textbf{0.181}&\textbf{0.023}&\textbf{4.856}&\textbf{0.996$\pm$0.002}\\
				cMLP&0.999&0.954&0.000&1.386&0.647$\pm$0.039\\
				cLSTM&0.854&1.000&0.000&1.674&0.522$\pm$0.053\\
				Pearson-Corr&0.844&0.979&0.004&1.960&0.793$\pm$0.060\\
				Kendall-Corr&0.889&0.980&0.002&1.957&0.798$\pm$0.038\\
				Spearman-Corr&0.865&0.981&0.003&1.957&0.801$\pm$0.042\\
				\hline			
			\end{tabular}
		\end{center}
	\end{table*}
	
	Table \ref{table4} shows that the proposed STDCDAE outperforms other methods in inferring the ECN under all different measures. Lower SSIM and CS prove that the ECNs inferred by STDCDAE for children and young adults are less similar. Higher JSD demonstrates that there are significant differences in the distribution of ROIs in the ECNs inferred by STDCDAE when comparing children to young adults. Meanwhile, the larger SED also suggests that the ECNs inferred by STDCDAE for different groups are distant in Euclidean space. The higher ACC indicates that the ECNs inferred by STDCDAE have significant differences and are easily distinguishable. The above results demonstrate that the ECNs inferred by STDCDAE can better characterize the differences between children and young adults, which derives further investigation to the changes in brain functional mechanisms as the brain develops.
	
	The results in Fig. \ref{fig2} and Table \ref{table4} demonstrate that STDCDAE is better in revealing the essential differences of brain functional networks, especially the differences of directed connections between children and young adults compared to alternative methods for the following reasons. On the one hand, both the dynamic deep encoder and deep decoder in STDCDAE adopt the spatio-temporal fusion mechanism to effectively utilize the spatio-temporal information contained in the rs-fMRI data. Furthermore, the causal constraints in learning criteria of STDCDAE focus the inferred ECNs on the most pivotal connectivity among ROIs. On the other hand, the ECNs inferred by STDCDAE are grounded in causality and account for information flow. This gives them an edge over FCNs, which solely rely on correlation.
	
	\subsection{Analysis of dECNs Representation}
\begin{figure*}[t]
		\includegraphics[width=\textwidth]{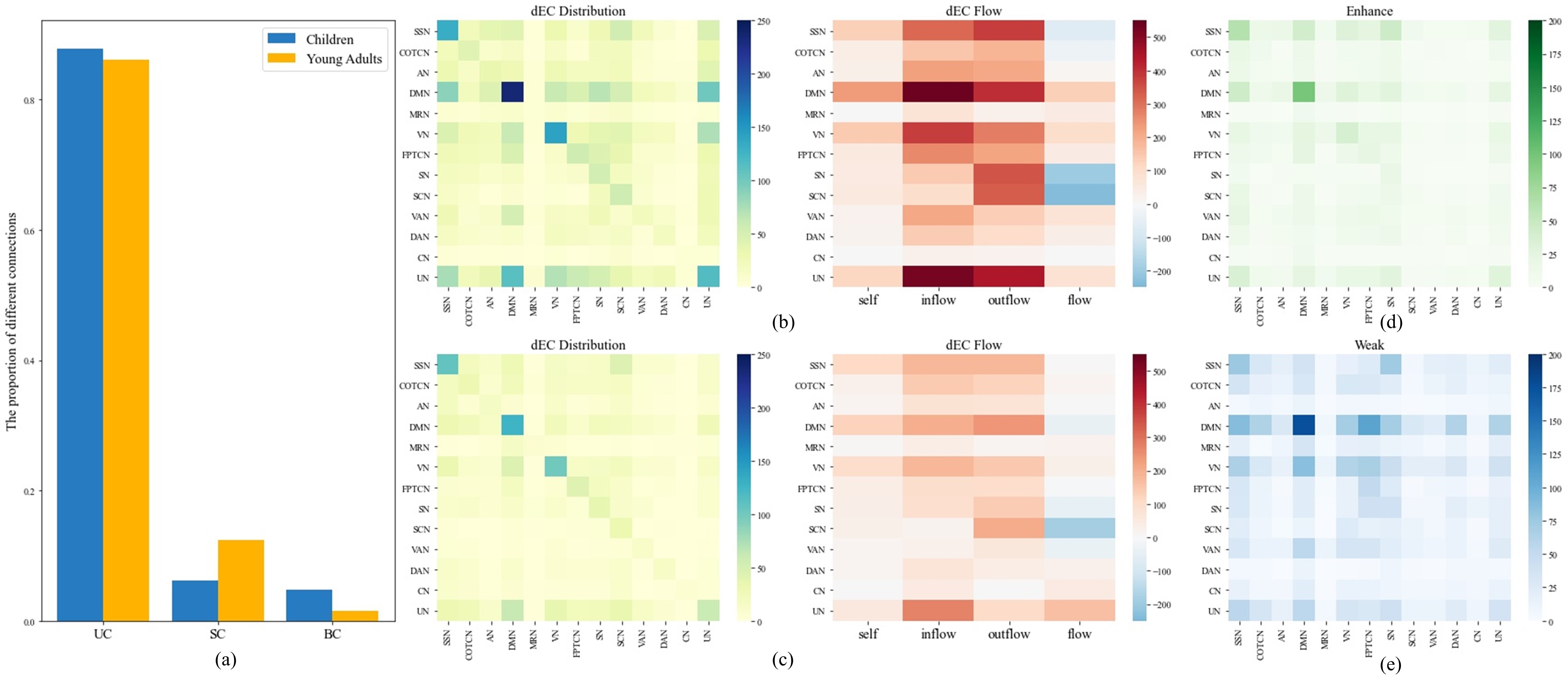}
		\caption{(a) The proportion of different connections between children and young adults. (b) The distribution of dECs in different RSNs and the  information interaction patterns of each RSN in children. (c) The distribution of dECs in different RSNs and the  information interaction patterns of each RSN in young adults. (d) The distribution of enhanced dECs in different RSNs. (e) The distribution of weakened dECs in different RSNs.}
		\label{fig3}
	\end{figure*}

	For the dECNs inferred by STDCDAE, the $t$-test is applied to detect the significantly nonzero dynamic effective connectivities (dECs), and the significantly nonzero dECs are defined as active dECs. There are 4017 active dECs in children and 2051 active dECs among young adults. The number of active dECs in young adults is smaller than that in children, indicating that there are many additional connections in the brain networks of children, which leads to more dECs in the dECNs during childhood. In order to further investigate the differences in dECN between children and young adults, the active dECs are further divided into unidirectional connectivity (UC), bidirectional connectivity (BC), and self-connectivity (SC). The definitions of UC, BC, and SC are as follows: $\text{dEC}_{ij}$ represents the dEC from the $j$-th ROI to the $i$-th ROI. When $i\ne j$, if $\text{dEC}_{ij}\ne 0$ and  $\text{dEC}_{ji}= 0$, then  $\text{dEC}_{ij}$ is defined as UC. If  $\text{dEC}_{ij}\ne 0$ and  $\text{dEC}_{ji}\ne 0$, then both $\text{dEC}_{ij}$ and $\text{dEC}_{ji}$ are defined as BC; When $i=j$, if $\text{dEC}_{ij}\ne 0$, it is defined as SC. According to the above definitions, there are 3573 UCs, 194 BCs, and 250 SCs in children, and 1767 UCs, 30 BCs, and 254 SCs in young adults. The proportions of UCs, BCs, and SCs in active dECs among different groups are shown in Fig. \ref{fig3} (a). Fig. \ref{fig3} (a) shows the proportion of UCs is much higher than the proportion of BCs and SCs. The proportion of BCs in children is significantly higher than that in young adults, but the proportion of SCs is significantly lower than that of young adults. This demonstrates that as age rises, the BCs between ROIs gradually degenerate into more efficient UCs, and SCs in some ROIs gradually form. This demonstrates that the function of ROIs gradually becomes specialized during brain development, and the functional structure of the brain becomes more efficient. This further confirms that, during brain development, the brain functional structure evolves from an undifferentiated framework to a distinct network.
	
	Compared to young adults, children's dECNs exhibit more redundant connections. As the brain matures, these redundant connections diminish—a finding consistent with \cite{stevens2009changes}. This is reflected in the fact that the number of active dECs in children is higher than that in young adults, and the proportion of bidirectional connectivities between ROIs in children is higher than that in young adults, and the proportion of self connections is lower than that in young adults. The distribution of active dEC in different brain networks shows that effective connections of children are more dispersed, while young adults are more concentrated, which is consistent with \cite{he2020decreased}. This conclusion further substantiates the idea that during brain development, the functional network evolves from a general system into a specialized one.
	
	In order to analyze the dECs between 264 ROIs from the perspective of brain function, the 264 ROIs are divided into 13 resting state networks (RSNs) \cite{power2011functional}, which contain the sensory/somatomotor network (SSN), cingulo-operational task control network (COTCN), frontal-parietal task control network (FPTCN), auditory network (AN), default mode network (DMN), memory retrieval network (MRN) visual network (VN), salience network (SN), subcortical network (SCN), ventral attention network (VAN), dorsal attention network (DAN), cerebellar network (CN), and uncertain network (UN). The first 12 RSNs are mainly related to motion, perception, cognition memory, language, and other brain functions, while the UN mainly includes 28 ROIs that are not strongly correlated with other RSNs. By comparing the  information interaction patterns within and between RSNs, there are significant differences in the  information interaction patterns between different RSNs of children and young adults, which demonstrates significant changes in the  information interaction patterns between different RSNs during brain development.
	
	Fig. \ref{fig3} (b) and (c) show the distribution of dECs in different RSNs and the  information interaction patterns of each RSN  for children and young adults respectively. The left in Fig. \ref{fig3} (b) and (c) shows that compared to young adults, the distribution pattern of brain functional networks in children is more dispersed, while young adults are more concentrated, which is consistent with the conclusion in \cite{yangExplainableMultimodalDeep2023}. There are lots of dECs in the SSN, DMN, and VN, where SSN is responsible for cognitive-related activities \cite{Londei2009SensorymotorBN}, DMN is responsible for internal psychological activities such as memory, internal thought, and imagination \cite{Raichle2001ADM}, and VN is responsible for processing visually information. These three RSNs not only have powerful internal information transmission but also have strong information inflow and outflow, which means that they are the brain functional hubs in resting state. However, compared to adults, children have stronger connectivity within DMN and between DMN and other RSNs, which reflects that brain functional networks of children are not yet specialized for processing information \cite{Cai2019CapturingDC}. Meanwhile, compared to children, the connectivity between the 12 RSNs with defined functions and UN in young adults is weaker, which reflects the gradual specialization of the brain functional network as the brain develops into adulthood. The right in Fig. \ref{fig3} (b) and (c) shows that there are large differences in the  information interaction patterns between children and young adults with different RSNs. Specifically, the inflow information of SSN in children is less than the outflow information, which is an information outflow network, but the inflow information of SSN in young adults is equal to the outflow information. In addition, COTCN is an information outflow network for children, but in young adults, the inflow information of COTCN is more than the outflow information. At the same time, DMN, FPTCN, VAN, which are information inflow networks RSNs in children, become the information outflow networks among young adults. In addition, compared to children, the inflow of information of CN in young adults has increased, which is different from the decline of inflow information of other RSNs with age, which may be led by the strong physical motor control ability of young adults. The above conclusions reflect that there are significant changes in the  information interaction patterns between different RSNs during brain development.
	
	\begin{figure}[h]
		\centering
		\includegraphics[width=3in]{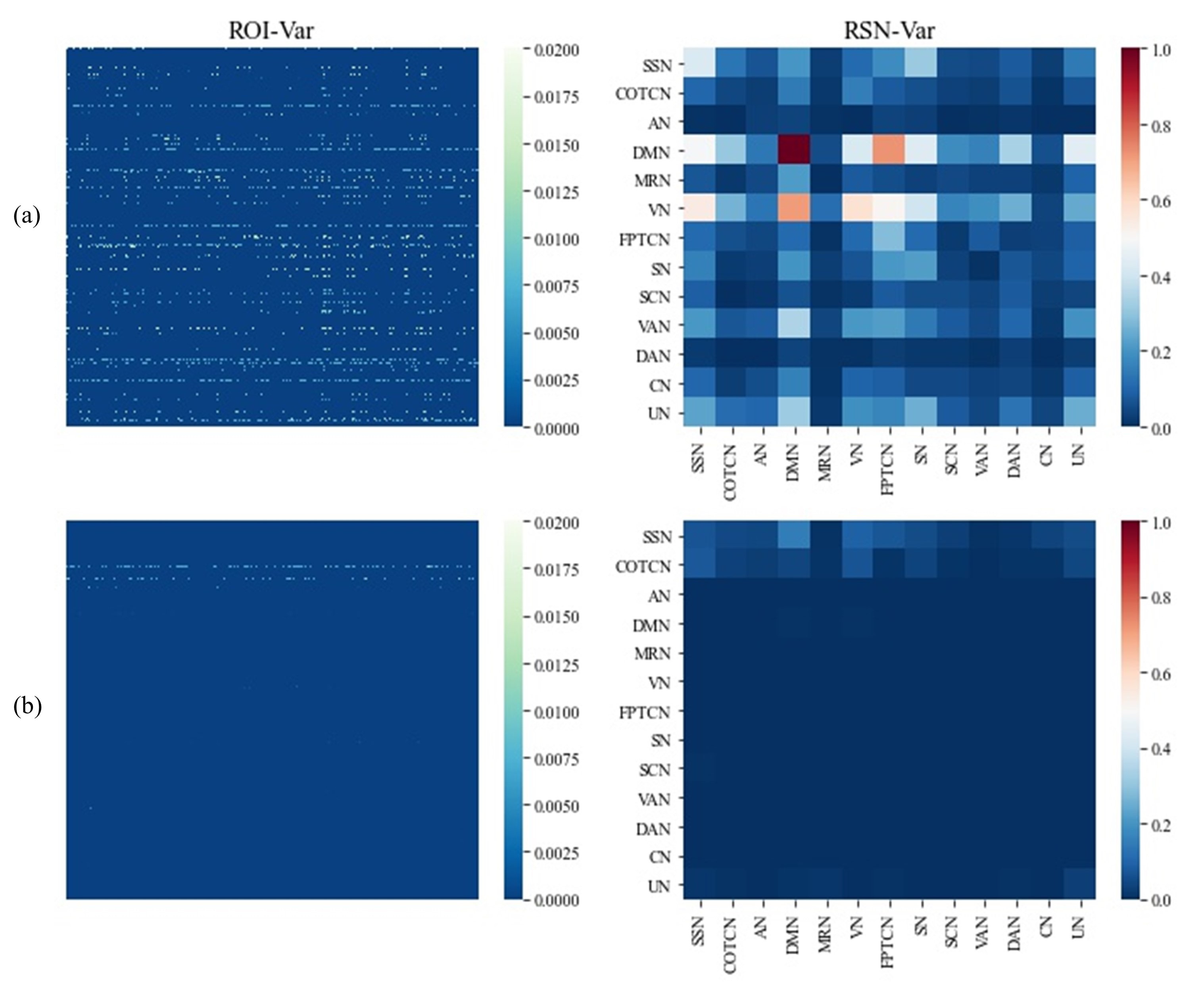}
		\caption{(a) The fluctuation of dEC intensity over time between ROIs and the fluctuation of dEC intensity within and between RSNs over time in children. (b) The fluctuation of dEC intensity over time between ROIs and the fluctuation of dEC intensity within and between RSNs over time in young adults.}
		\label{fig4}
	\end{figure}
	
	In order to further investigate the changes in dECs with age, the hypothesis testing methods in \cite{Qiao2021ADA} is used to determine dEC with significant differences. Through the hypothesis testing methods, 3569 significantly weakened dECs and 1708 significantly enhanced dECs are found among the 6068 active dECs. The number of enhanced dECs is smaller than the number of weakened dECs which reflects the redundant connectivities in the brain network of children gradually disappear during brain development, and the brain functional network pattern becomes more integrated with age. Fig. \ref{fig3} (d) shows that compared to children, young adults have stronger connectivity within  SSN, DMN, and between SSN, DMN, and other RSNs (such as VN, FPTCN, and SN), which is consistent with \cite{Cai2018EstimationOD}. The increased connectivity between VN, SSN, and DMN confirms that there are some super-connectivities between VN and these RSNs \cite{Kelly2009DevelopmentOA}, which indicates that young adults frequently have higher visual information processing ability than children. Fig. \ref{fig3} (e) shows that there are many weakened dECs within or between SSN, COTCN, DMN, VN, FPTCN, SN, and VAN. The many weakened dECs within and between RSNs demonstrate that as the brain develops and complicated cognition gradually forms, the extraneous connectivities in the brain network of children gradually disappear. The weakened dECs between DMN and FPTCN reflect a higher level of cognition among young adults compared to children because the strength of connectivity between DMN and FPTCN is negatively correlated with higher cognitive performance \cite{Feng2019VerbalCI}. The weakened dECs between DMN and SN also explain why children can respond to external threats faster than adults, as the interaction between DMN and SN is an important factor in controlling cognition. The connectivities of these two networks are associated with internal psychological activities, attention, and motor function \cite{Bonnelle2012SalienceNI}.
	
	Through the analysis of dECs with significant differences between children and young adults, we find that dECs with significant differences are primarily distributed within or between RSNs such as SSN, COTCN, DMN, VN, FPTCN, SN, and VAN. These RSNs are associated with information processing, memory, cognition, sensory and motor functions. There are both weakened dECs and enhanced dECs within DMN and between DMN and other RSNs, demonstrating broad and complicated connectivities between DMN and other RSNs, which is consistent with \cite{Cai2018EstimationOD}. Meanwhile, the enhanced dECs between SSN, DMN, and VN indicate that young adults have higher visual information processing abilities than children. The weakened dECs between DMN and FPTCN reflects that young adults have higher cognitive levels compared to children. The weakened connectivity between DMN and SN reveals the reason why children can respond faster to external threats compared to adults. In addition, the number of weakened dECs is more than the number of enhanced dECs, which confirms that compared to young adults, the brain functional networks of children are more dispersed, and as age increases, the brain functional network shifts from inter-regional dispersion to intra-regional integration.
	
	\subsection{Analysis of dECNs Time-varying Patterns}
	\begin{figure*}[!h]
		\includegraphics[width=\linewidth]{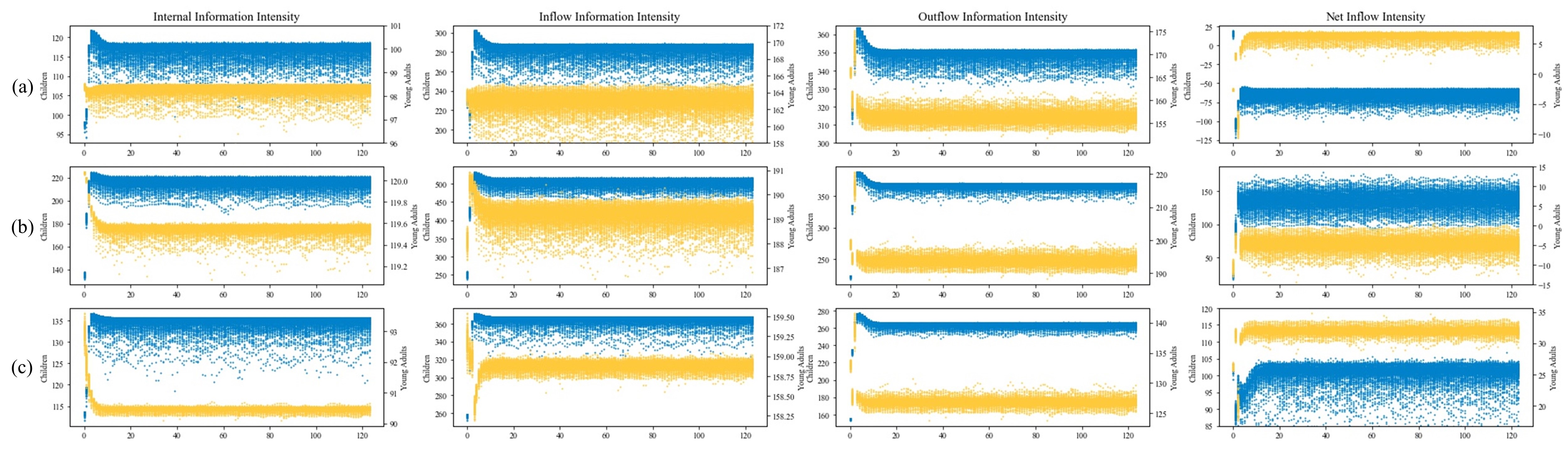}
		
		\includegraphics[width=\linewidth]{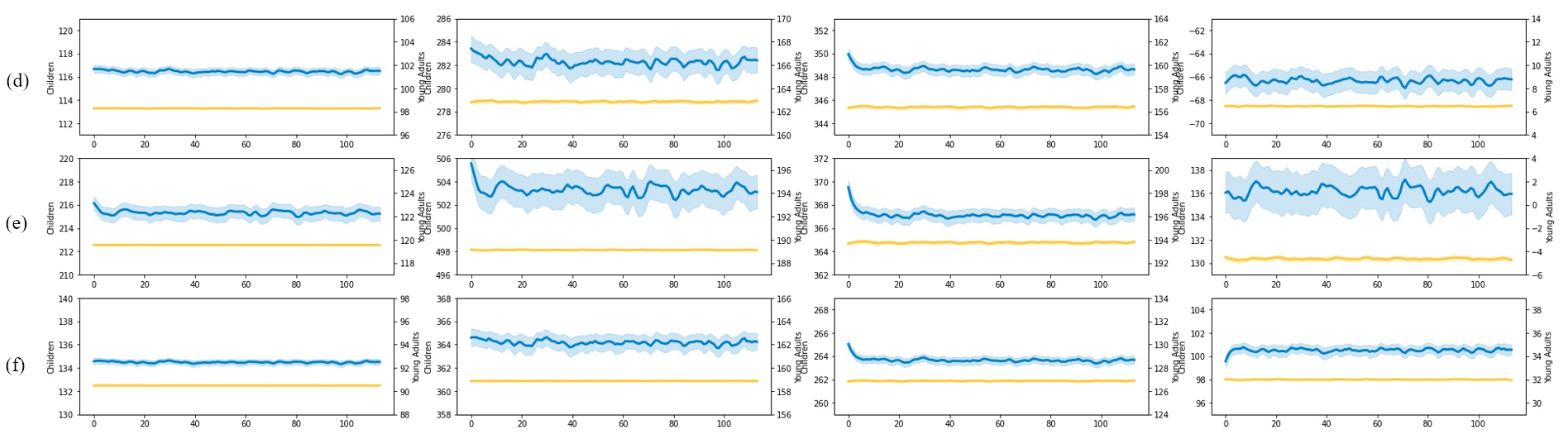}
		\caption{The changes in information exchange patterns of brain functional network hubs over time. In this figure, the x-axis represents scan time and the y-axis represents the information flow intensity, every point stands for the directed flow intensity between RSNs of a subject at that time. (a) Changes in SSN information interaction patterns for children and young adults over time, with blue representing children and yellow representing young adults (the same below). The mean and 95\% confidence interval are shown in (d). (b) Changes in DMN information interaction patterns for children and young adults over time. The mean and 95\% confidence interval are shown in (e). (c) Changes in VN information interaction patterns for children and young adults over time. The mean and 95\% confidence interval are shown in (f).}
		\label{fig5}
	\end{figure*}
	
	To comprehensively analyze the differences in dECN between children and young adults, we examined the time-varying patterns of dECNs across groups. The variance of dECs intensity over time is applied for measuring the fluctuation of dEC intensity between ROIs, and the fluctuation of dEC intensity within and between RSNs. Fig. \ref{fig4} (a) and (b) show the fluctuation of dEC intensity over time between ROIs, as well as the fluctuation of dEC intensity within and between RSNs over time, for children and young adults respectively. The Fig. \ref{fig4} illustrates that compared to young adults, there is a stronger fluctuation in dEC between various ROIs in children. Meanwhile, the number of dECs with significant fluctuations in children is more than that of young adults, and  dECs with significant fluctuation are found in various RSNs in children. This reflects that compared to children, the brain functional network of young adults is more stable. This also indicates that children are more sensitive. In addition, there are significant fluctuations in dECs between SSN and COTCN within them, as well as between RSNs other than MRN, in both children and young adults, because SSN is related to cognitive activities, and COTCN is related to information transmission between the RSNs, they both remain active in rest state \cite{sheffield2015fronto}. At the same time, the dECs within and between MRNs in both children and young adults are very stable with almost no fluctuations. This stability might be attributed to the role of MRN in encoding, storing, and retrieving memory, especially since this region remains inactive during the resting state \cite{sestieri2011episodic}. Based on the fluctuation of connectivity intensity among RSNs of children, we find that there are significant fluctuations within DMN, VN, and between them and other RSNs, especially within DMN, in children. The significant fluctuation between DMN and other RSNs in children confirms the stronger connectivity between DMN and other RSNs as previously mentioned. Combining with the Fig. \ref{fig3} (e), the reason that young adults have smaller fluctuations within and between RSNs than children may be the weakening of connectivity strength. This also reflects the disappearance of redundant connections in the brain functional network during brain development, which not only makes the brain functional network more integrated and improves information processing efficiency but also enhances the stability of the brain functional network.
	
	The brain functional network hubs, SSN, DMN, and VN are chosen to further analyze the changes in their  information interaction patterns over time, and investigate the differences in dECNs between different groups. Meanwhile, in order to avoid the effects of asynchronous in rs-fMRI between children and young adults, we mainly focus on the overall trends and common characteristics of each group. Fig. \ref{fig5} (a), (b) and (c) show the changes in the internal information flow intensity, the inflow information flow intensity, the outflow information flow intensity, and the net inflow intensity (which is the difference between inflow intensity and outflow intensity) over time for SSN, DMN, and VN of children and young adults, where blue represents children and yellow represents young adults. Fig. \ref{fig5} shows that the three functional network hubs only fluctuate briefly at the start due to the T1 effect of fMRI, and then become stable quickly due to the adaptation of brain \cite{HSU2004270}. In order to examine the fluctuation between children and young adults during stable period, the first ten time points are ignored. Using the same measuring scale, the mean and 95\% confidence interval of Fig. \ref{fig5} (a), (b) and (c) are computed and are shown in Fig. \ref{fig5} (d), (e) and (f). After becoming stable, the internal information flow intensity, inflow information flow intensity and outflow information flow intensity of children's hubs are higher than those of young adults, which further verifies that the brain function network of children is more active than young adults. Meanwhile, the SSN, which is an information outflow network in children, becomes an information inflow network in young adults. Similarly, the DMN changes from an inflow network in children to an outflow one in young adults, signifying a shift in the information interaction patterns between SSN, DMN, and other networks. In addition, the direction of the information flow in the three brain functional network hubs remained unchanged after entering a stable state.
	
	By analyzing the time-varying patterns of dECNs, it is evident that young adults possess a more stable brain functional network connectivity pattern than children. This stability may arise from the elimination of extraneous connections and diminished connectivity strength. After entering the stable state, the internal information flow intensity, inflow information flow intensity and outflow information flow intensity of hubs of children are higher than those of young adults, which further verifies that directed brain function networks of children are more active than young adults. In addition, the  information interaction patterns between DMN, SSN, and other RSNs have undergone significant changes in brain development.
	
	\section{Conclusion}
	In this article, we propose a model named STDCDAE to reveal dynamic causality in brain activity data. By utilizing spatio-temporal information in the fMRI data, STDCDAE can accurately discover dynamic causality. Specifically, a dynamic causal deep encoder leverages spatio-temporal information to identify causality, while a decoder verifies its accuracy. Experimental results from both linear and nonlinear simulation data validate the accuracy of STDCDAE in identifying causality. Experimental results from real fMRI data in PNC show that the dECNs inferred by STDCDAE can better characterize the differences between children and young adults. Through the analysis of inferred dECNs, we find that the directed information flow and interaction patterns between RSNs undergo significant changes during development, implying the gradual maturation of the human brain. In addition, the proposed model highlights the critical role of dECNs in uncovering the dynamics of effective information processing and adaptability within the brain, demonstrating the use of deep learning with good explainability and robustness.

	\section*{Acknowledgments}
	This research was supported by the National Natural Science Foundation of China (Nos. 12090021, 12271429 and 12226007), the National Key Research and Development Program of China (No. 2020AAA0106302), the Natural Science Basic Research Program of Shaanxi (No. 2022JM-005), the Science and Technology Innovation Plan of Xi'an (No. 2019421315KYPT004JC006) and was partly supported by the National Institutes of Health (R01 MH104680, R01 GM109068, R01 MH121101, R01 MH116782, R01 MH118013 and P20-GM144641) and the HPC Platform, Xi'an Jiaotong University.

	\bibliographystyle{elsarticle-num}
	\bibliography{STDCDAE}
	
\end{document}